\def\year{2017}\relax
\documentclass[letterpaper]{article}
\usepackage{aaai17}
\usepackage{times}
\usepackage{helvet}
\usepackage{courier}

\usepackage{times}
\usepackage{epsfig}
\usepackage{graphicx}
\usepackage{amsmath}
\usepackage{amssymb}
\usepackage{algorithm}
\usepackage{algorithmic}
\usepackage{multirow}
\usepackage{subfigure}
\usepackage{times}
\usepackage{float}

\begin{document}
 \setlength\titlebox{2.5in}
\title{Ordinal Constrained Binary Code Learning for Nearest Neighbor Search}
\author{
Hong Liu$^{\dag\ddag}$,
Rongrong Ji$^{\dag\ddag *}$,
Yongjian Wu$^{\natural}$,
Feiyue Huang$^{\natural}$ \\
$^{\dag}$Fujian Key Laboratory of Sensing and Computing for Smart City, Xiamen University, 361005, China\\
$^{\ddag}$School of Information Science and Engineering, Xiamen University, 361005, China\\
$^{\natural}$BestImage Lab, Tencent Technology (Shanghai) Co.,Ltd, China\\
\texttt{lynnliu.xmu@gmail.com, rrji@xmu.edu.cn, littlekenwu@tencent.com, garyhuang@tencent.com} \\
}
\maketitle
\begin{abstract}
Recent years have witnessed extensive attention in binary code learning, \textit{a.k.a.} hashing, for nearest neighbor search problems.
It has been seen that high-dimensional data points can be quantized into binary codes to give an efficient similarity approximation via Hamming distance.
Among existing schemes, ranking-based hashing is recent promising that targets at preserving ordinal relations of ranking in the Hamming space to minimize retrieval loss.
However, the size of the ranking tuples, which shows the ordinal relations, is quadratic or cubic to the size of training samples.
By given a large-scale training data set, it is very expensive to embed such ranking tuples in binary code learning.
Besides, it remains a dificulty to build ranking tuples efficiently for most ranking-preserving hashing, which are deployed over an ordinal graph-based setting.
To handle these problems, we propose a novel ranking-preserving hashing method, dubbed \textit{Ordinal Constraint Hashing} (OCH), which efficiently learns the optimal hashing functions with a graph-based approximation to embed the ordinal relations.
The core idea is to reduce the size of ordinal graph with ordinal constraint projection, which preserves the ordinal relations through a small data set (such as clusters or random samples).
In particular, to learn such hash functions effectively, we further relax the discrete constraints and design a specific stochastic gradient decent algorithm for optimization.
Experimental results on three large-scale visual search benchmark datasets, \textit{i.e.} LabelMe, Tiny100K and GIST1M, show that the proposed OCH method can achieve superior performance over the state-of-the-arts approaches.
\end{abstract}

%%%%%%%%%%%%%%%%%%%%%%%%%%%%%%%%%%%%%%%%%%%%%%%%%%%%%%%%%%%%%%%
\section{Introduction}
Learning binary code, \textit{a.k.a.} hashing,  to preserve the data similarity has recently been popular in various computer vision and artificial intelligence applications, \emph{e.g.}, image retrieval \cite{liu2016towards}, objective detection \cite{Dean2013FastAD}, multi-task learning \cite{Weinberger2009FeatureHF}, linear classifier training \cite{Li2011HashingAF,Lin2014LibD3CEC}, and active learning \cite{Liu2012CompactHH}.
In this setting, real-valued data points are encoded into binary codes that are significantly efficient in storage and computation.
In general, most hashing methods learn a set of hash functions $h^k: \mathbf{R}^d \rightarrow \{0,1\}^r$.
It typically maps the $d$-dimensional data space into an $r$-bit discrete Hamming space, such that the nearest neighbors can be approximated by using the compact binary codes learned.

Recent advances in binary code learning can be categorized into either data-independent or data-dependent ones \cite{Wang2016LearningTH}.
The former typically refers to random projection/partition of feature space, such as Locality Sensitive Hashing (LSH) and Min-Hash (MinHash).
It typically requires long bits or multi-hash table to achieve satisfied retrieval performance.
Both supervised and unsupervised hashing belong to data-dependent hashing.
Unsupervised hashing, learns hash functions by preserving the data structure, distribution, or topological information, \emph{e.g.},  Spectral Hashing (SH) \cite{Weiss2008SpectralH}, Anchor Graph Hashing (AGH) \cite{liu2011hashing}, Isotropic Hashing (IsoHash) \cite{kong2012isotropic}, Iterative Quantization (ITQ) \cite{gong2013iterative}, Discrete Graph Hashing (DGH) \cite{Liu2014DiscreteGH}, Spherical Hashing (SpH) \cite{heo2015spherical}, Scalable Graph Hashing (SGH) \cite{jiang2015scalable}, and Ordinal Embedding Hashing (OEH) \cite{liu2016towards}.
Differently, supervised hashing aims to learn more accurate hash  functions with label information.
Representative works include, but not limited to, Binary Reconstructive Embedding (BRE) \cite{Kulis2009LearningTH}, Minimal Loss Hashing (MLH) \cite{Norouzi2011MinimalLH}, Kernel-based Supervised Hashing (KSH) \cite{liu2012supervised}, Semi-Supervised Hashing (SSH) \cite{Wang2012SemiSupervisedHF}, Supervised Discrete Hashing (SDH) \cite{Shen2015SupervisedDH}.
%Despite significant accuracy gain, it is typically labour-intensive to obtain sufficient semantic labels in many real-world applications.

%Although promising performance has been shown from these methods, we argue that,  the relative order between data pairs, rather than the pairwise distance, which must be preserved in the Hamming space.
Although promising performance has been shown from these methods, we argue that, the relative order among data must be preserved in the Hamming space rather than pairwise relations.
So, many ranking-based hashing algorithms have been proposed to learn more discriminative hash codes, \textit{e.g.}, Hamming Distance Metric Learning (HDML) \cite{Norouzi2012HammingDM}, Ranking-based Supervised  Hashing (RSH) \cite{Wang2013LearningHC}, Struct-based Hashing (StructHash) \cite{Lin2014OptimizingRM}, Top-Rank Supervised Binary Coding (Top-RSBC) \cite{Song2015TopRS}.
However, most of these methods adopt the stochastic gradient decreasing (SGD) optimization under triplet ordinal constraints, which needs massive iterations.
On the other hand, such ranking-based hashing are categorized into supervised hashing, which is typically labor-intensive to obtain sufficient semantic labels in many real-world applications.

In this paper, we mainly focus on ranking preserved hashing, termed Ordinal Constraint Hashing (OCH), with two key innovations to address the issues raised above.
\begin{figure}[!t]\label{fig1}
\begin{center}
\includegraphics[height=0.7\linewidth,width=1.0\linewidth]{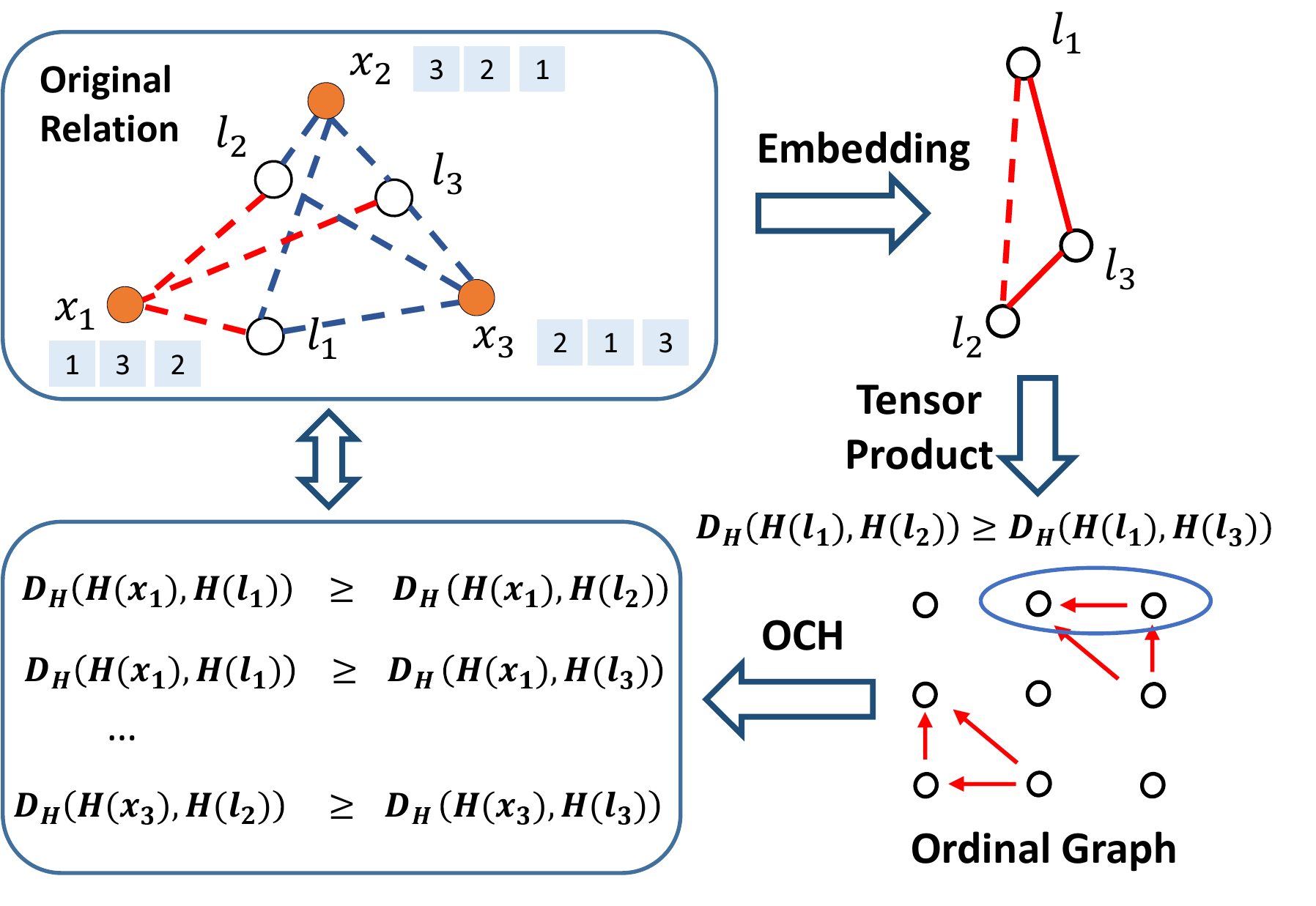}
\setlength{\abovecaptionskip}{-8pt}
\caption{ The Framework of our proposed Ordinal Constraint Hashing (OCH).}
\end{center}
\end{figure}
Firstly, OCH  attempts to preserve the ordinal relation from the high-dimensional feature space to the Hamming space.
Our learning procedure takes into account the \emph{quartic ordinal relations}, rather than the pairwise or triplet ordinal relations that are widely used in the earlier works.
To make use of the quartic relations, the ranking lists for each individual queries are converted to a quartic tuples representation that can be presented as Eq. 1 below.
Formally, let $x$ be the data point, a quartic tuples can be given as:
%\small{
\begin{equation} \label{eq1}
\mathbb{J}\! =\! \{ (q;x_i,x_j,x_k) | D(q,x_i)\! < \!D(q,x_j)  \! < \!D(q,x_k) \},
\end{equation}
where $D(\cdot,\cdot)$ is the distance measure (\textit{e.g.} Euclidean distance).
To the best of our knowledge, none existing hashing methods, even ranking-based hashing \cite{li2013learning,wang2013learning,liu2016towards}, have considered such ordinal relation among data points, which can be obtained in a large-scale manner with extensive human labor.

Inspired by the recent work \cite{babenko2016pairwise}, OCH embeds in which the original quartic order relation can hold as the triplet order relation.
And the size of order tuples can be reduced from the original $[n^4]$ to $[L^3]$ ($n << L$) to avoid high complexity computation cost, where $n$ is the number of training data and $L$ is the number of the sub-set (\textit{e.g.} clusters).
Then, we propose a general method to construct the ordinal graph with tensor product calculation.
Such construction scheme avoids the time consuming selection way to represent the  ordinal tuples.
The OCH minimizes the inconsistency between the given ordinal relation tuples and the ones derived from the corresponding hash codes.
At last, due to the constraint of orthogonal projection, the hash functions are learned via a special SGD algorithm, which formulate the problem as combing the traditional SGD with Stiefel manifold optimization.
The whole framework is shown in Fig. 1.
We compare the proposed OCH against various state-of-the-art unsupervised hashing methods on three widely used similarity search benchmarks, \textit{i.e.}, \textbf{LabelMe}, \textbf{Tiny100K}, and \textbf{GIST1M}.
Quantitative experiments demonstrate that OCH outperforms the existing unsupervised hashing methods in terms of both accuracy and efficiency.

The rest of this paper is organized as follows:
In Section 2, we briefly overview the related works of the proposed method.
Section 3 and 4 describe the proposed OCH and the iterative SGD based optimization.
In Section 5, we show  and analyze the experimental results.
Finally, we conclude this paper in Section 6.
%%%%%%%%%%%%%%%%%%%%%%%%%%%%%%%%%%%%%%%%%%%%%%%%%%%%%%%%%%%%%%%
\section{Background and Related Work}
We briefly introduce the problem of binary code learning and review related work as below:

\textbf{Binary Code Learning} aims to learn a set of hash functions to encode real-valued feature points to compact binary codes.
For a data point $x \in \mathbb{R}^d$, the hash functions $H = \{ h_1, ..., h_r \}$ produces a $r$-bit binary code $y = \{ y_1, y_2, ..., y_r\}$ for $x$ as:
\begin{equation} \label{eq2}
y = [ h_1(x), ..., h_k(x), ..., h_r(x) ].
\end{equation}
Therefore, the $k$-th hash bit $y_k$ is calculated by $h_k(x) = sgn\big(f_k(x)\big)$, where $sgn(\cdot)$ is the sign function that returns 1 if $f_k(x)>0$ and -1 otherwise.
Such hash functions are encoded as a mapping process combining with quantization, which has been widely used in many traditional hashing algorithm, \textit{e.g.}, LSH.
And the function $f_k: \mathbb{R}^d \to \mathbb{R}$ is a linear transformation, given by $f_k(x) = W_k^Tx+b$ with the projection matrix $W_k$ and an offset $b\in\mathbb{R}$.
Then given a set of hash functions, the database $X \in\mathbb{R}^{d \times n}$ with $n$ samples are mapped to the produced binary codes as
\begin{equation} \label{eq3}
Y = \big\{ h_1(X), ..., h_k(X),..., h_r(X) \big\},
\end{equation}
where $Y \in \{0,1\}^{r\times n}$ is the hash code matrix of the database $X$.

\textbf{Ordinal Embedding Hashing} \cite{liu2016towards} aims to learn a set of hash functions as Eq. \ref{eq3}.
Such functions can preserve the ordinal relations between $\delta_{ij}$ and $\delta_{kl}$, where $\delta_{ij}$ is the dissimilarity between the $i$-th and the $j$-th item.
Its goal is to make sure the ordinal relation can be preserved in the produced Hamming space: $\delta_{ij} < \delta_{kl}: \left\Vert H(x_{i})-H(x_{j})\right\Vert_1 <\left\Vert H(x_{k})-H(x_{l})\right\Vert_1.$
To embed such ordinal relations, OEH first constructs a directed unweighted ordinal graph $G=(V,E) = [n^4]$, where each node $v_{ij}$ is the dissimilar degree $\delta_{ij}$, and each directed edge is defined via $e_{(i,j,k,l)}=(v_{ij}\rightarrow v_{kl})\subseteq E$.
Then the objective function is to minimize the inconsistency between the given ordinal relation graph and the ones generated from the corresponding hash codes.
At last, by using the landmark-based ordinal graph, the quartic ordinal relation is transformed to the triplet ordinal relation, which transforms the target of OEH to $\delta_{ij} < \delta_{ik}: \left\Vert H(x_{i})-H(l_{j})\right\Vert_1 <\left\Vert H(x_{i})-H(l_{k})\right\Vert_1$ where $l_{j}$ and $l_{k}$ are the landmark points.
%%%%%%%%%%%%%%%%%%%%%%%%%%%%%%%%%%%%%%%%%%%%%%%%%%%%%%%%%%%%%%%
\section{Ordinal Constraint Hashing}
In this section, we describe the proposed OCH in detail.
%First, we introduce some notations before describing the proposed hashing algorithm.
Let $X = \{ x_1, x_2, ..., x_n \} \in \mathbb{R}^{d\times n}$ be the data matrix with $n$ samples, where  $x_i$ is the $i$-th column of $X$ with $d$ dimensions.\footnote{Without loss of generality, assume that the data $X$ is normalized and mean-centered.}
As the same definition as $\delta_{ij}$ before, we suppose that $\delta_{ii} = 0$ and $\delta_{ij} = \delta_{ji}$, the comparison $\delta_{ij}<\delta_{kl}$ reflects the data pair $(i,j)$ is more similar than pair $(k,l)$.
Then an oridinal relation set $C$ can be given as: $\{\delta_{ij} < \delta_{ik} < \delta_{il} ~|~ \forall (i;j,k,l) \in C\}$.
We further define a K-means centers points matrix $L=\{l_1,l_2,...,l_L\}\in \mathbb{R}^{d\times L}$, where $L << n$.

The proposed OCH aims to learn the hash functions by embedding the ordinal relation.
To this end, a straightforward method is to maximize the loss function between the ordinal relation set $C$ and the corresponding relation in the Hamming space.
This can be defined by the following objective function:
\begin{equation}\label{eq4}
\setlength\jot{0pt}
\begin{split}
&\begin{split}
& max\sum_{(i;j,k,l)\in C}I\big(D_{H}(b_{i},b_{j})\leq D_{H}(b_{i},b_{k})\leq D_{H}(b_{i},b_{l})\big) \\
\end{split}\\
&\begin{split}
 s.t. \: ~~ &b_{i}=sgn(W^{T}x_{i}), ~~ W^TW=I,~~ W\in \mathbb{R}^{d\times r},
\end{split}
\end{split}
\end{equation}
where $I(\cdot)$ is an indicator function which returns 1 if the condition is satisfied and 0 otherwise, and $D_H(b_i,b_j)$ returns the Hamming distance between hash code $b_i$ and $b_j$.

\begin{figure}[!t]\label{fig2}
\begin{center}
\includegraphics[height=0.34\linewidth]{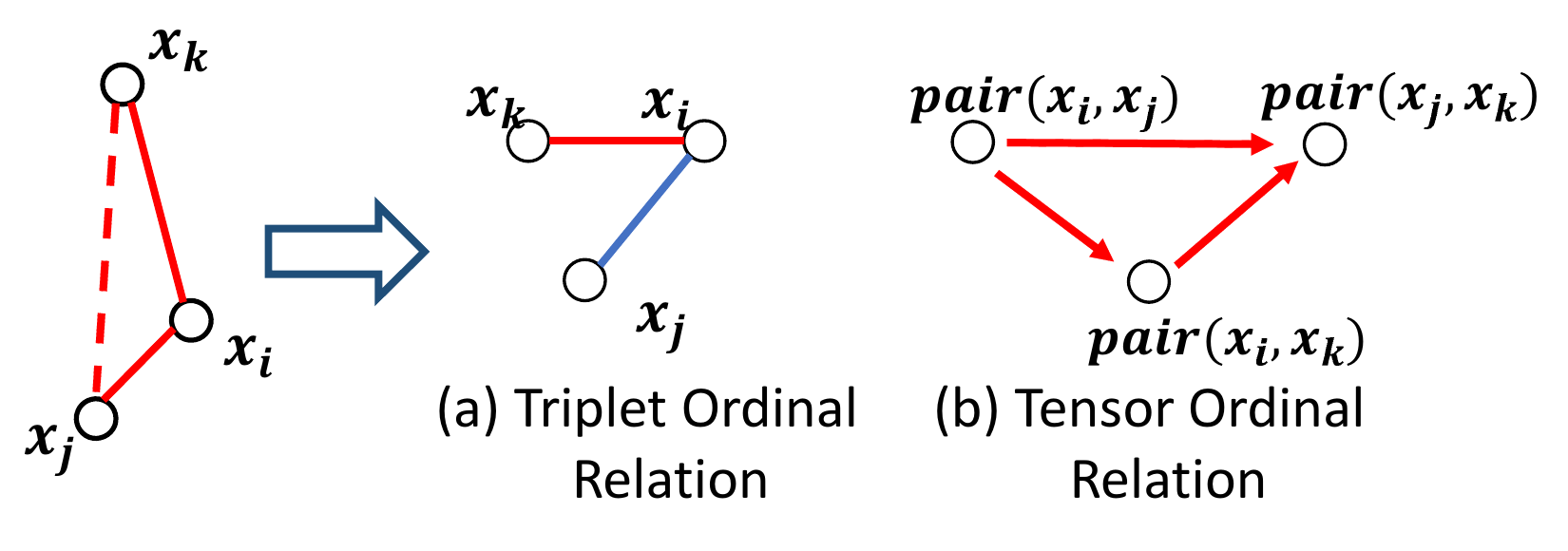}
\setlength{\abovecaptionskip}{-8pt}
\caption{ The ordinal relation for a training data.}
\end{center}
\end{figure}

%\subsection{Ordinal Graph with Tensor Product}

The first key problem of OCH is how to represent the ordinal relation set directly.
In the previous works, the ordinal relation is represented with the triplet data $(x_i,x_j,x_k)$, where the pair $(x_i,x_j)$ is formed by two nearest neighbors under Euclidean distance, and $(x_i,x_k)$ is dissimilar pair.
This ordinal relation is shown in Fig. 2 (a), where red line represent the dissimilar pair and blue line represent the similar pair.
However, this triplet representation needs to randomly select the corresponding triplet tuples and compare all the data points side by side, which time consuming and memory costly.
Even, it is hard to define the similar and dissimilar data pairs for unsupervised learning problems.

In this paper, we address the above issue by proposing a novel ordinal graph model with tensor product calculation, in which triplet or even quartic ordinal relation can be efficiently represented.
Given a dataset $X$ and the similarity measure (\textit{e.g.} Euclidean Distance), the affinity graph $S\in \mathbb{R}^{n \times n}$ is constructed as follows:
\begin{equation} \label{eq5}
    S(i,j)=  \begin{cases}
        0, & i=j , \\
        e^{-\left \| x_i - x_j \right \|^2_2/2 \sigma^2 }, & otherwise,
    \end{cases}
\end{equation}
where $S(i,j)$ represents the similarity between two data points.
We further define a dissimilarity graph $DS \in \mathbb{R}^{n\times n}$, with each entry as $DS(i,j) = 1 / S(i,j)$ and $DS(i,i) = 0$ (which is the definition of $\delta_{ij}$).

Then, a tensor ordinal graph (TOG) $\mathbf{G}$ is defined as:
\begin{equation} \label{eq6}
    \mathbf{G} = S \otimes DS,
\end{equation}
where $ \otimes $ is the Kronecker product of matrices defined as $\mathbf{G}(ij,kl) = S(i,j) \cdot DS(k,l)$.
Thus, each entry of $\mathbf{G}$ relates to four data points.
When $S$ and $DS$ are two $n \times n$ matrices, $\mathbf{G}$ is an $n^2 \times n^2$ matrix.
Therefore, the ordinal relation between the quartic items in $(i,j,k,l) \in C$ can be represented through the TOG, as following:
\begin{equation} \label{eq7}
    \begin{cases}
        \delta_{ij} < \delta_{kl}, & \mathbf{G}(ij,kl) > 1 , \\
        \delta_{ij} > \delta_{kl}, & \mathbf{G}(ij,kl) \leq 1.
    \end{cases}
\end{equation}

Fig.2 (b) shows a toy example.
Under such a circumstance, the ordinal relation in original feature space is $\delta_{ij} < \delta_{ik} < \delta_{jk}$, which can be simply calculated and compared with Euclidean distance.
And for the quartic item $(i,k,i,j)$, the $(ik,ij)$-th entry is $G(ik,ij) = S(i,k) \cdot DS(i,j) = S(i,k) / S(i,j)$.
Due to the relation of $S(i,k) < S(i,j)$, we can get $\mathbf{G}(ik,ij)<1$, which reflects the truth ordinal relation  $\delta_{ik}>\delta_{ij}$ in the original space.
In such way,  the proposed TOG can represent the ordinal relation simply with tensor product scheme.

%Although the ordinal relations can be generated easily, the relations in set $C$ should also be calculated and compared by using TOG, which is not what we expect.
%Although the TOG approximates the relation set $C$ easily, the larger size of TOG makes the calculation time consuming, not what we expect.
Although the TOG approximates the relation set $C$ easily, it is not expect that the larger size of TOG makes the calculation time consuming.
To solve this problem, according to the relation in set $C$, we transform the constraint $\{\delta_{ij} < \delta_{ik} < \delta_{il} ~|~ \forall (i;j,k,l) \in C\}$ as follows:
\begin{equation} \label{eq8}
\begin{split}
    \mathbf{O} = \sum_{\forall (i\neq j,k,l)} I \big( (\| x_i - &x_j \|^2_2 - \| x_i - x_k \|^2_2)^2 \\[-3mm]
&     - (\| x_i - x_j \|^2_2 - \| x_i - x_l \|^2_2)^2 \big).
\end{split}
\end{equation}
That is to say, minimizing Eq. \ref{eq8} is to approximate the ordinal constraint set $C$ over all dataset.
Since the points in the dataset have been normalized, we rewrite Eq. \ref{eq8} as:
\begin{equation} \label{eq9}
        \begin{split}
     \mathbf{O}  = &  \sum_{\forall (i\neq j,k,l)}\! I \big( (x_i^Tx_j - x_i^Tx_k)^2 - (x_i^Tx_j - x_i^Tx_l)^2 \big) \\
    & \begin{split}
              =   \sum_{\forall (j,k,l)}\! I \big( (x_j-x_l&)^TM(x_j-x_l) \\[-3mm]
              &-  (x_j-x_k)^TM(x_j-x_k)  \big),\\
     \end{split}
        \end{split}
\end{equation}
where $M = \sum_{i} x_i^Tx_i$ is a positive semi-definite symmetrical matrix.
So it is convenient to use SVD to decompose it into $Z\in \mathbb{R}^{d_{svd} \times d}$ such that $M = Z^T\Lambda Z$.
Then a mapping function can be defined as $u_i = Zx_i \in \mathbb{R}^{d_{svd}}$, and Eq. \ref{eq9} is written as following:
\begin{equation} \label{eq10}
\begin{split}
    & \begin{split}
    \mathbf{O} = \sum_{\forall (j,k,l)} I \big((u_j-&u_l)^T\Lambda(u_j-u_l) \\[-3mm]
   &- (u_j-u_k)^T\Lambda (u_j-u_k) \big)
    \end{split}\\[1mm]
               & \leq \sum_{\forall (j,k,l)} I \big( \|\Lambda^{\frac{1}{2}} \|^2_2 \cdot (\| u_j - u_l \|^2_2- \| u_j - u_k \|^2_2) \big) \\
               & \propto \sum_{\forall (j,k,l)} I \big( \| u_j - u_l \|^2_2- \| u_j - u_k \|^2_2 \big).
\end{split}
\end{equation}
By means of this mapping, termed ordinal constraint projection (OCP), we have projected the original ordinal relation to an approximation ordinal relation set $\{\hat{\delta}_{ij} < \hat{\delta}_{ik} ~|~ \forall (i,j,i,k) \in \hat{C}\}$, which can be generated from the TOG easily.

On the other hand, the total number of ordinal constraints is still too large to be used for training.
Inspired by the Product Quantization \cite{jegou2011product}, the distance $D(u_i, u_j)$ can be approximated by the distance $D(u_i,u_j) \approx D(a_i, a_j)$, where $a_i = Zl_i$ is the  embedding center point.
In our setting, we further approximate the original ordinal set $C$ by a sub-set of ordinal relation set after the aforementioned OCP.
Then the size of the tensor ordinal graph can be reduced to $[L^4]~(L<<n)$,  where $L$ is the number of K-means centers.
From Eq. \ref{eq10}, we can use the triplet relation among centers for the original quartic relation  approximation, which significantly reduce the scale of ordinal graph from $[n^4]$ to  $[L^3]$.
%From Eq. \ref{eq10}, we can use a sub-set of the TOG for approximation, which significantly reduce the scale of ordinal graph from $[n^4]$ to  $[L^3]$.

Therefore, the overall objective function in Eq. \ref{eq4} for the proposed OCH approach is rewritted as follows:
\begin{equation}\label{eq11}
\setlength\jot{0pt}
\begin{split}
&\begin{split}
& min\sum_{(i,j,i,k)\in \hat{C}}I\big(D_{H}(b_{i},b_{j})\geq D_{H}(b_{i},b_{k}) \big) \\
\end{split}\\
&\begin{split}
 s.t. \:  &b_{i}=sgn(V^{T}a_{i}), ~~VV^{T}=I,~~ V\in \mathbb{R}^{d_{svd}\times r}.
\end{split}
\end{split}
\end{equation}
Due to $W^TW=I$ with $W = Z^TV$, we can easily get the new orthogonal constraint in Eq. \ref{eq11}, which needs to be hold during optimization.
Meanwhile our target is to learning the hash functions that hold the ordinal relations in the Hamming space.
As a result, a new optimization scheme should be designed, which is introduced  subsequently in Section 4.

\section{Optimization}

Directly minimizing the objective function in Eq. \ref{eq11} is intractable,
as the coding function is discrete while the Hamming space is not continuous.
To solve this problem, we relax the discrete
constraints from the Hamming space to an approximated continuous space.

To that effect, we first relax the hashing function $H(a_i)=sgn(V^{T}a_i)$
as follows:
\begin{equation}\label{e12}
\hat{H}(a_i)=tanh(V^{T}a_i),
\end{equation}
where $tanh(\cdot)$ is a good approximation for $sign(\cdot)$ that transforms the binary codes  from $\{0,1\}$ to $\{-1,1\}$.
Correspondingly, the Hamming distance is calculated as:
\begin{equation}\label{e13}
{D}_{H}(b_{i},b_{j})=\frac{1}{2}\big(r-\hat{H}^{T}(a_i)\cdot\hat{H}(a_j)\big).
\end{equation}
Finally, we use the Sigmoid function to replace the indicator function for convenient optimization and avoid overfitting.
Based upon the above relaxations, the objective function in Eq. \ref{eq11} can be rewritten as:
\begin{equation}\label{e14}
{F}(V|G) = \sum_{(i,j,i,k)\in \hat{C}} p(i,j,i,k),~~ s.t.~~ VV^{T}=I,\\
\end{equation}
where the $p(\cdot, \cdot, \cdot, \cdot)$ is the Sigmoid function defined as follows:
\[
p(i,j,i,k) = \frac{1}{1+exp\big(D_H(b_i,b_k)-D_H(b_i,b_j)\big)}.
\]
Intuitively, the gradient descent approach can be used to carry out an
iterative optimization for Eq. \ref{e14}.
However, due to the orthogonal constraints of projection matrix $V$, the objective function is non-convex, which is also hard to optimize.
In the following, we further introduce an alternative stochastic gradient descent algorithm on Stiefel Manifold to solve this problem efficiently.

\begin{algorithm}[t]
\caption{Ordinal Embedding Hashing (OEH)}
\renewcommand{\algorithmicrequire}{\textbf{Input:}}
\renewcommand{\algorithmicensure}{\textbf{Output:}}
\begin{algorithmic}[1]
\REQUIRE
Data set $X=\{x_1, x_2, ..., x_n\}$, parameters $\gamma$ and $\eta$.
\ENSURE
The hash function $H(x_i) = sgn(V^TZx_i)$.
\STATE Generate centers $L$ by K-means algorithm;
\STATE Generate matrix $Z$ and embed $Z$ into the $L$;
\STATE Generate ordinal relations set $\hat{C}$ by Tensor Ordinal Graph in Eq. \ref{eq7};
\REPEAT
\STATE
Randomly select a subset $c$ from set $\hat{C}$;
\STATE
Calculate the gradient according to  Eq. \ref{e16};
\STATE
Update $V$ according to Eq. \ref{e15};
\UNTIL{convergence or reaching the maximum iteration number.}
\end{algorithmic}
\end{algorithm}

% \usepackage{multirow}
\begin{table*}[t]
\centering
\caption{The \emph{m}AP and Precision Comparison Using Hamming Ranking on Two Benchmark with Different Hash Bits}
\label{my-label}
\scalebox{0.85}[0.85]{
\begin{tabular}{|c|c|c|c|c|c|c|c|c|c|c|c|c|}
\hline
\multirow{3}{*}{Methods} & \multicolumn{6}{c|}{LabelMe}                            & \multicolumn{6}{c|}{Tiny100K}                           \\ \cline{2-13}
                         & \multicolumn{3}{c|}{mAP} & \multicolumn{3}{c|}{Pre@100} & \multicolumn{3}{c|}{mAP} & \multicolumn{3}{c|}{Pre@100} \\ \cline{2-13}
                         & 32     & 64     & 128    & 32      & 64      & 128      & 32     & 64     & 128    & 32      & 64      & 128      \\ \hline
LSH                      & 0.1582 & 0.2563 & 0.3555 & 0.2894  & 0.4233  & 0.5543   & 0.1231 & 0.1831 & 0.2323 & 0.2524  & 0.3845  & 0.4626   \\ \hline
AGH                      & 0.2099 & 0.2349 & 0.2297 & 0.3895  & 0.4424  & 0.4730   & 0.1176 & 0.1350 & 0.1221 & 0.4286  & 0.3751  & 0.3687   \\ \hline
IsoHash                  & 0.2606 & 0.3225 & 0.3883 & 0.4486  & 0.5211  & 0.5939   & 0.1864 & 0.2295 & 0.2599 & 0.4107  & 0.4906  & 0.5064   \\ \hline
SpH                      & 0.2481 & 0.3096 & 0.3859 & 0.4476  & 0.5244  & 0.6232   & 0.1931 & 0.2541 & 0.3193 & 0.4404  & 0.5599  & \textbf{0.6755}   \\ \hline
SGH                      & 0.3012 & 0.3849 & 0.4477 & 0.4899  & 0.5920  & 0.6589   & 0.2054 & 0.2637 & 0.3038 & 0.4444  & 0.5380  & 0.5969   \\ \hline
ITQ                      & 0.3059 & 0.3753 & 0.4210 & 0.5012  & 0.5724  & 0.6176   & 0.2039 & 0.2436 & 0.2612 & 0.3044  & 0.4912  & 0.5164   \\ \hline
OEH                      & 0.2111 & 0.3546 & 0.4449 & 0.3693  & 0.5642  & 0.6500   & 0.1650 & 0.2495 & 0.3086 & 0.3635  & 0.4986  & 0.5954   \\ \hline
OCH                      & \textbf{0.3140} & \textbf{0.3947} & \textbf{0.4620} & \textbf{0.5072}  & \textbf{0.6028}  & \textbf{0.6713}   & \textbf{0.2297} & \textbf{0.2919} & \textbf{0.3379} & \textbf{0.4805}  & \textbf{0.5785}  & 0.6314   \\ \hline
\end{tabular}}
\end{table*}

\subsection{Stochastic Gradient Descent on Stiefel Manifold}

Optimization with respect to the orthogonal constraints has been recently studied in \cite{Wen2013AFM,Ge2014OptimizedPQ}.
In particular, to solve Eq. 14, most straightforward method uses  gradient descent on the Stiefel manifold defined by $\mathcal{O} = \{V\in \mathbb{R}^{d_{svd}\times r}, VV^T=I \}$ \cite{absil2009optimization}.
An off-the-shelf iterative solver has been developed in \cite{Wen2013AFM}, which is however sensitive to the initialization and hard to integrated into our optimization.

Recent advances in \cite{cunningham2015linear} are to calculated the objective $F$, gradients $\nabla F$ together in the full space $\mathbb{R}^{d_{svd}\times r}$.
These gradients are then projected into a tangent space $\mathbb{T}$ with the transformation $\mathbf{P}:\mathbb{R}\rightarrow \mathbb{T}$, and a retraction $\mathbf{R}:\mathbb{T}\rightarrow \mathcal{O}$ is adopted to map the gradients from tangent space to the targeted Stiefel manifold space\footnote{Both two transformation functions have been defined in the appendix of \cite{cunningham2015linear}.}.
Therefore, this generic algorithm offers a global convergence proof for such a method by a line search \cite{absil2009optimization}.
Finally, the updating rule of the optimal projection matrix can be defined as:
\begin{equation} \label{e15}
    V = \mathbf{R}\big(\eta \mathbf{P}(-\nabla F)\big)
\end{equation}
where $\eta$ is the choice of convergence parameter, which is widely used for first-order optimization.
In this way, the gradient of Eq. \ref{e14} in the full space is given by:
\begin{equation}\label{e16}
\begin{split}
	\nabla F& =\\
	& \sum_{c\subset \hat{C}_s} \!{\big(p(c)(1\!-\!p(c))\big)\cdot}\! \left[\frac{\partial{D}_{H}(b_{i},b_{k})}{\partial V}\!-\!\frac{\partial{D}_{H}(b_{i},b_{j})}{\partial V}\right],
\end{split}
\end{equation}
where $I$ is an identity matrix, $c$ is a subset random selected from the whole ordinal relations $\hat{C}$, and the gradient of Hamming distance is formulated as:
\begin{equation}\label{e17}
\begin{split}
\frac{\partial{D}_{H}(b_{i},b_{j})}{\partial V}= -\frac{1}{2}\big\{a_i &\cdot\left[\left(\text{1-\ensuremath{\hat{H}^{2}}(\ensuremath{a_i})}\right)\odot\hat{H}(a_j)\right]^{T} \\
&+ a_j\cdot\left[\left(\text{1-\ensuremath{\hat{H}^{2}}(\ensuremath{a_j})}\right)\odot\hat{H}(a_i)\right]^{T}\big\}.
\end{split}
\end{equation}
In Eq. (\ref{e17}), $\odot$ is the Hadamard product which represents the element-wise
product.

The details of the proposed SGD on Stiefel manifold is shown in Algorithm 1.
The overall training complexity of the proposed algorithm is $O(trL^3d_{svd} + nL)$, where $t$ is the number of iterations.
It is linear to the training set and related to the complexity of the K-means step, which is faster than the previous work \cite{liu2016towards} in ranking preserved hashing.
The experiments shown in the next section prove  that the  proposed OCH has superior  performance for large-scale similarity retrieval with high efficiency in training.

%%%%%%%%%%%%%%%%%%%%%%%%%%%%%%%%%%%%%%%%%%%%%%%%%%%%%%%%%%%%%%%
\section{Experiments}
In this section, we evaluate the proposed Ordinal Constraint Hashing on three  large-scale   benchmarks, \textit{i.e.}, LableMe, Tiny100K, and GIST1M, which are widely used for evaluating nearest neighbor search algorithms.
\subsection{Datasets}
We briefly summarize the datasets used as below:
The \textbf{LabelMe} dataset consists of $22,019$ images, each of which is represented by a 512-dimensional GIST feature \cite{oliva2001modeling}.
The \textbf{Tiny-100K-384D} dataset consists of $100K$ images sampled from the TinyImages dataset  \cite{torralba2008small}, each of which is represented by a $384$-dimensional GIST descriptors.
The \textbf{GIST-1M-960D} dataset is introduced in  \cite{jegou2011product}, which consists of one million images described by GIST descriptors.

\subsection{Baseline Methods}
We compared the proposed OCH with several representative and state-of-the-art  unsupervised hashing methods, including Local Sensitive Hashing (\textbf{LSH}) \cite{datar2004locality},  Anchor Graph Hashing (\textbf{AGH}) \cite{liu2011hashing}, Isotropic Hashing (\textbf{IsoHash}) \cite{kong2012isotropic}, Iterative Quantization (\textbf{ITQ}) \cite{gong2013iterative}, Spherical Hashing (\textbf{SpH}) \cite{heo2015spherical}, Scalable Graph Hashing (\textbf{SGH}) \cite{jiang2015scalable}, and Ordinal Embedding Hashing (\textbf{OEH}) \cite{liu2016towards}.\footnote{The source codes of all the above methods are provided by authors kindly.}
%All the above are unsupervised methods including the proposed one.
For all the compared methods, we carefully follow the original parameter setting  in respective datasets.
We implement our OCH hashing using MATLAB on a single PC with Duo-Core I7-3421 and 75G memory, where the complete data set can be stored.

\begin{figure*}[!th]
\begin{center}
\begin{minipage}[t]{0.3\linewidth}
\centerline{
\subfigure[\emph{m}AP curves on GIST1M.]{
\includegraphics[width=\linewidth]{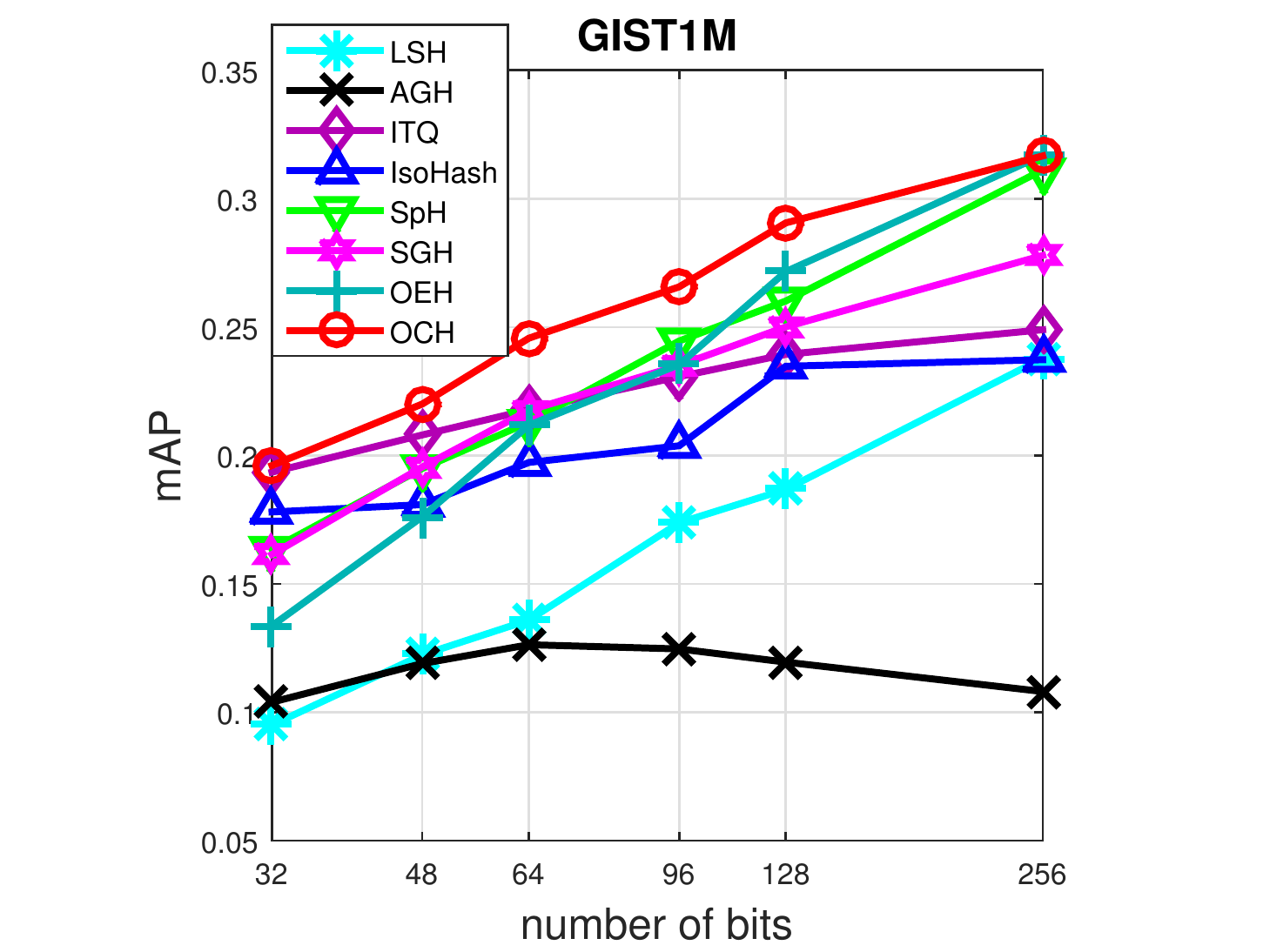}}\hspace*{-0.2\linewidth}
\subfigure[Rec$@K$ curves on GIST1M.]{
\includegraphics[width=\linewidth]{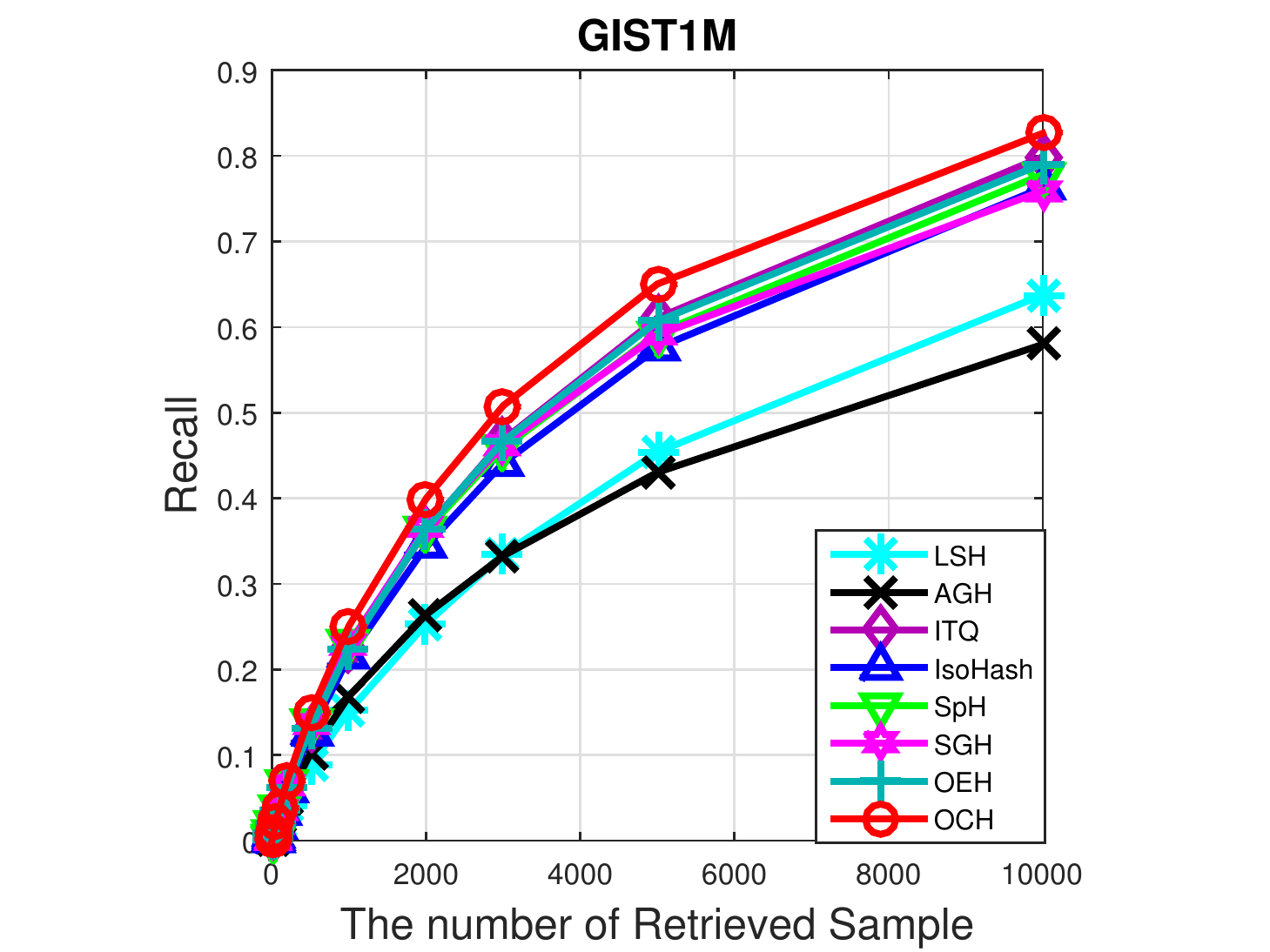}}\hspace*{-0.2\linewidth}
\subfigure[Rec$@K$ on LabelMe.]{
\includegraphics[width=\linewidth]{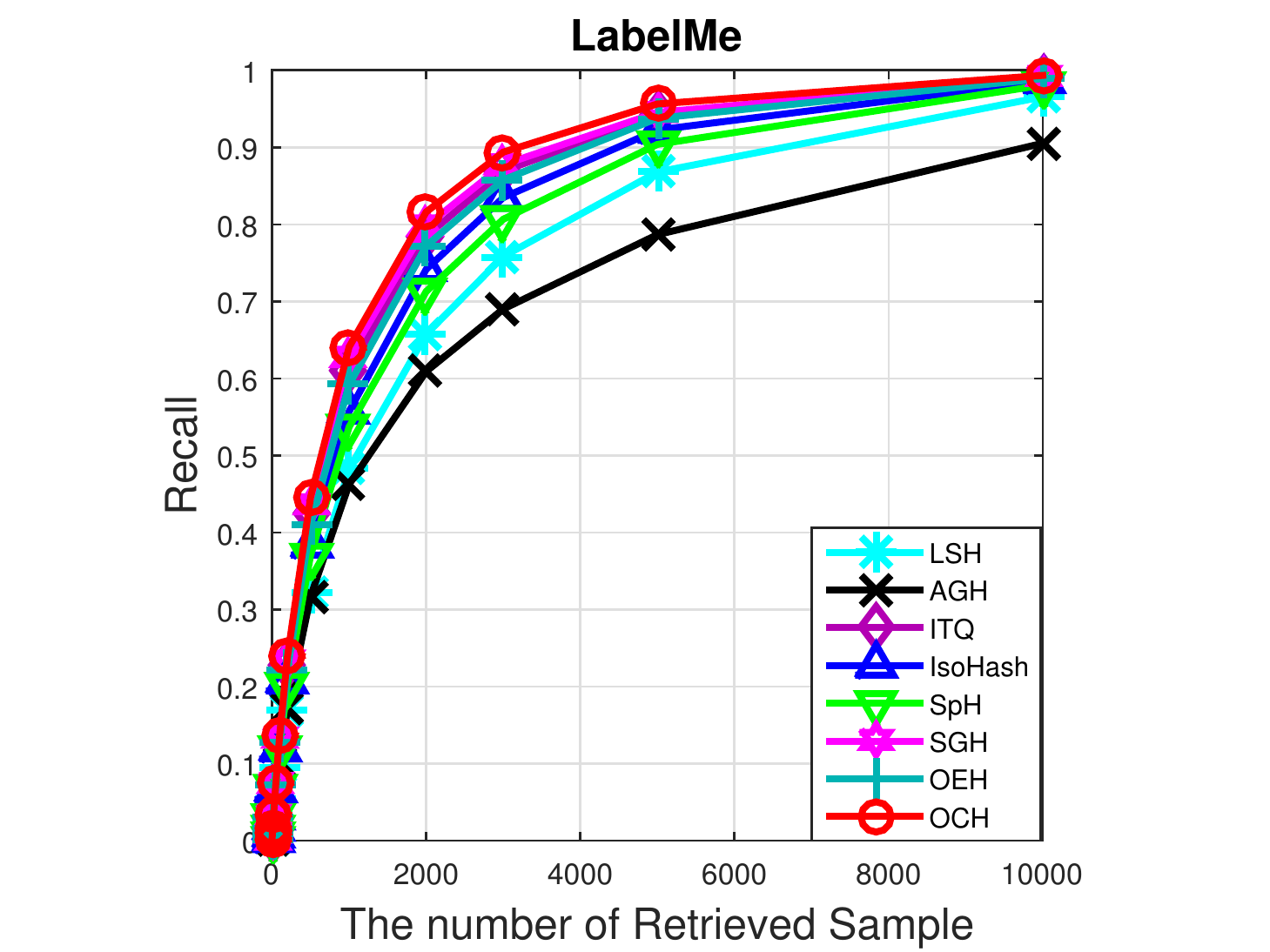}}\hspace*{-0.2\linewidth}
\subfigure[Rec$@K$ on Tiny100K.]{
\includegraphics[width=\linewidth]{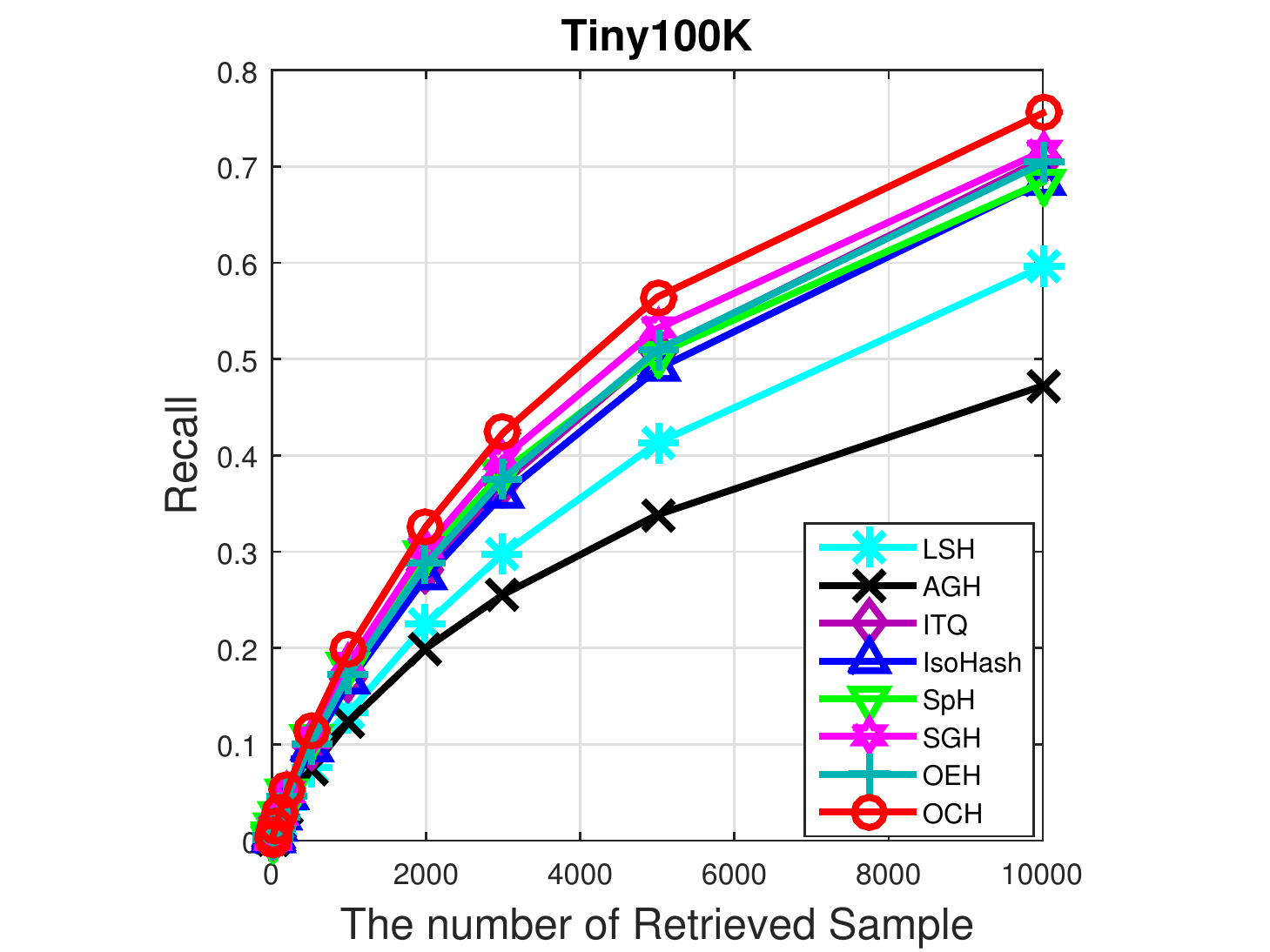}}
}
%\centerline{\small (2) MIR-Flickr25K Dataset}
\end{minipage}
\end{center}
\setlength{\abovecaptionskip}{-8pt}
\caption{The \emph{m}AP and Recall Curves of all the algorithms on three benchmarks. ( (b)-(d) are all evaluated on 64 bit.) }
\end{figure*}

\subsection{Evaluation Protocols}
To evaluate the proposed hashing algorithm, we adopt a set of widely used protocols in recent papers \cite{jiang2015scalable,park2015neighbor,liu2016towards}.
For a given query, the top $2\%$ ranking items with Euclidean distances are defined as with the same label of the query.
Then, based on the Euclidean ground-truth, we compute the recall curve, precision$@100$ ranking curve, and mean average precision.
For all the experiments, $2,000$ data points are randomly selected as test set, and the remaining are used as the dataset.
To avoid the overfitting, we randomly select $10,000$ points as the training set for all the algorithms, which has been used in \cite{he2013k}.
We run all the experiments $10$ times and report the average performance.

\subsection{Parameter Tunning}
In particular, $V$ is randomly initialized with a Gaussian distribution of mean $0$ and standard deviation $1$, which follows the standard settings.
%
%The learning rate $\eta$ is set as $0.1$ in all experiments.
%
For the constraints in Eq. \ref{eq11}, for each dataset the centers are formed of
 $300$ points  obtained by K-means clustering.
We  also give experimental analysis on whether the number of centers affects the retrieval performance in Fig. 4 (a).
It is worth to note that centers generated by K-means reflect the distribution and structure of data points, which can approximate the original ordinal graph without performance reduction.
For the SVD dimension, we set $16$ to the parameter $d_{svd}$, which can get the competitive performance by reducing data noise.
The number of centers generated by K-means Clustering is set to $300$, and the corresponding analysis will be given in below section.
%Following \cite{liu2016towards}, we set the parameter $\beta$ as $1$ in all our experiments to achieve the best search precision.

\subsection{Quantitative Results}
%We compare the proposed OCH with the state-of-the-art unsupervised hashing methods on three different scale datasets: \textbf{LableMe}, \textbf{Tiny100K} and \textbf{GIST1M}.

As shown in Tab. 1 and Fig. 3 with hash code varied from $32$ to $128$, the  proposed OCH consistently achieves superior performance over all baselines among all datasets, especially when the hash bit is short.

\begin{table}[t]
\centering
\caption{The Training Time (s) comparing with different algorithms on both datasets.}
\label{my-label2}
% \hspace*{-0.1\linewidth}
\scalebox{0.9}[0.9]{
\begin{tabular}{|c|c|c|c|c|c|c|}
\hline
              & \multicolumn{2}{c|}{LabelMe} & \multicolumn{2}{c|}{Tiny100K} & \multicolumn{2}{c|}{GIST1M} \\ \hline
Methods       & 32         & 64         & 32          & 64          & 32            & 64            \\ \hline
LSH           & 0.01       & 0.01       & 0.04        & 0.04        & 0.01          & 0.01          \\ \hline
ITQ           & 0.47       & 1.02       & 0.46        & 0.78        & 0.64          & 0.97       \\ \hline
SGH           & 1.61       & 3.25       & 1.62        & 3.21        & 1.81          & 4.43        \\ \hline
OEH           & 49.00      & 83.79      & 48.38       & 80.90       & 62.2    4         & 115.53         \\ \hline
\textbf{OCH}  & 29.07      & 35.01      & 28.89       & 35.93       & 29.11         & 36.69         \\ \hline
\end{tabular}}
\end{table}

Most previous hashing works always quantize the hash code by minimizing the loss between Euclidean distance and Hamming distance.
But in OCH, we change to minimize the inconsistency between original loss  before and after learning binary codes.
Compared to the previous works, it is quantitatively demonstrated that preserving such ordinal cues in hashing is a more fundamental goal for nearest neighbor search.
As for the comparison between our work and the most recent works in ranking preserve hashing OEH \cite{liu2016towards}, in comparison, both OEH and OCH adopt a two-step projection to find the optimal binary quantization space.
For OEH, the PCA dimension reduction is used as the first step, which cannot reduce the scale of training  and  still needs about $nL^2$ triplet order tuples to train the overall model.
On the contrary for the proposed OCH, we use the ordinal constraint projection in the first step, which not only reduces the original feature dimension, but also reduces the scale of training.
Tab. 2 further shows the comparison of training time between the above methods, in which OCH costs about half training time of OEH but achieves better performance on all the three datasets.
Moreover, OEH needs to construct the ordinal relations during each iteration, which is time consuming.
In contrast, our OCH constructs the ordinal relations by TOG only once before iteration, which is very convenient and efficient in iterative training.

At last, we discuss the influence of centers and the number of relations used in each iteration.
Note that the centers are generated by K-means, and the final \emph{m}AP result in Fig. 4 (a) is shown with the number of centers increasing from $100$ to $800$ among three datasets.
We find that the performances rarely change with the number increasing  of centers.
As a result, only a small number of centers is used  to approximate the overall ordinal relations, which is set to $300$ for all the datasets.
As in  Fig. 4 (b), we show the relation between \emph{m}AP and the training time by increasing the number $k(L-k)$ of randomly selected ordinal relations  during each iteration.
The result shows that the training time is linear to the number of ordinal  relations, but the \emph{m}AP does not change too much.
Therefore, we can use a small set of ordinal relations in each iteration, while maintaining the overall search precision.

\begin{figure}[t]
\begin{center}
\begin{minipage}[t]{0.5\linewidth}
\centerline{
\subfigure[\emph{m}AP vs. $\#$centers.]{
\includegraphics[width=\linewidth]{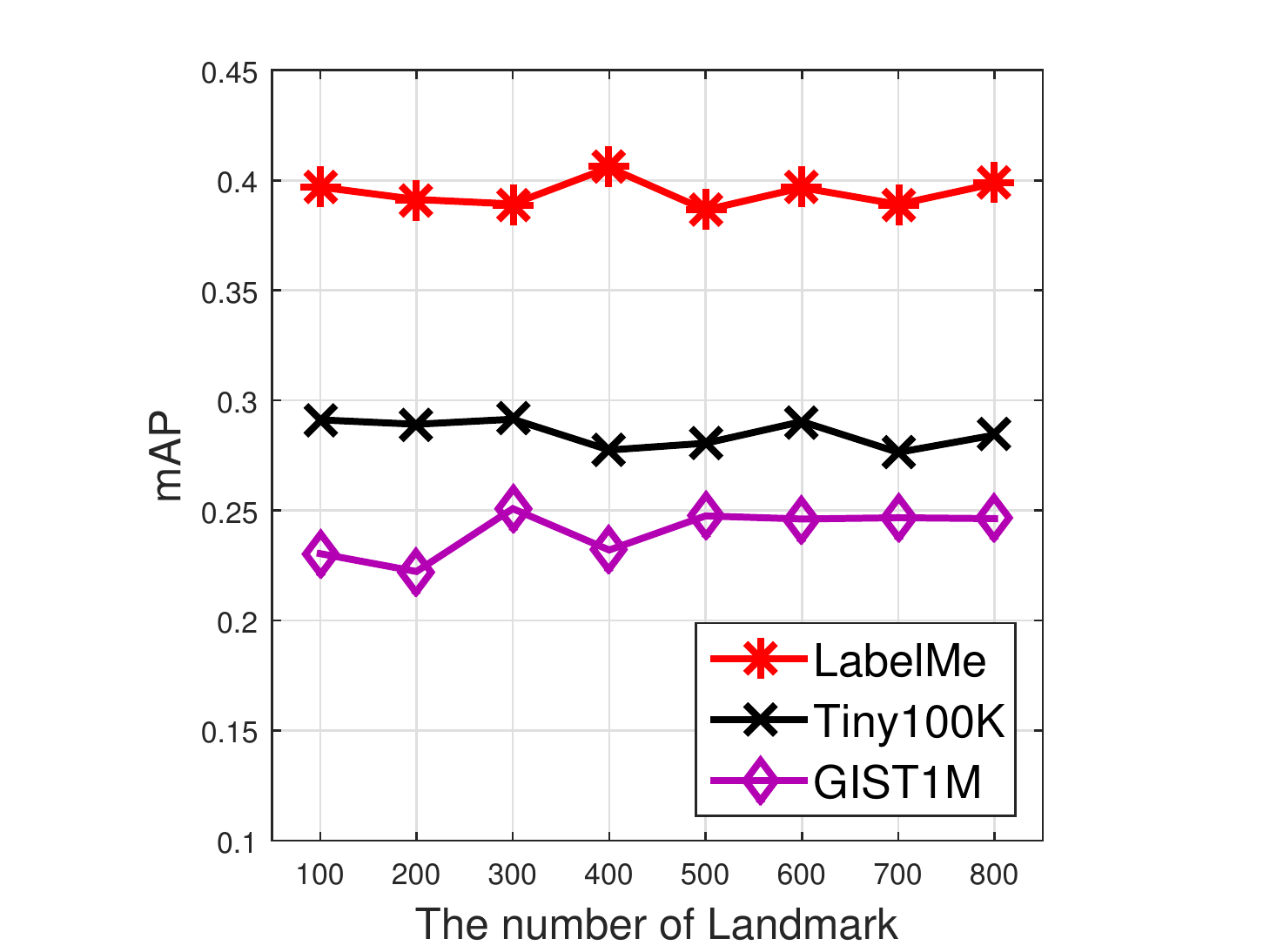}}\hspace*{-0.2\linewidth}
\subfigure[\emph{m}AP vs. the number of subset.]{
\includegraphics[width=\linewidth]{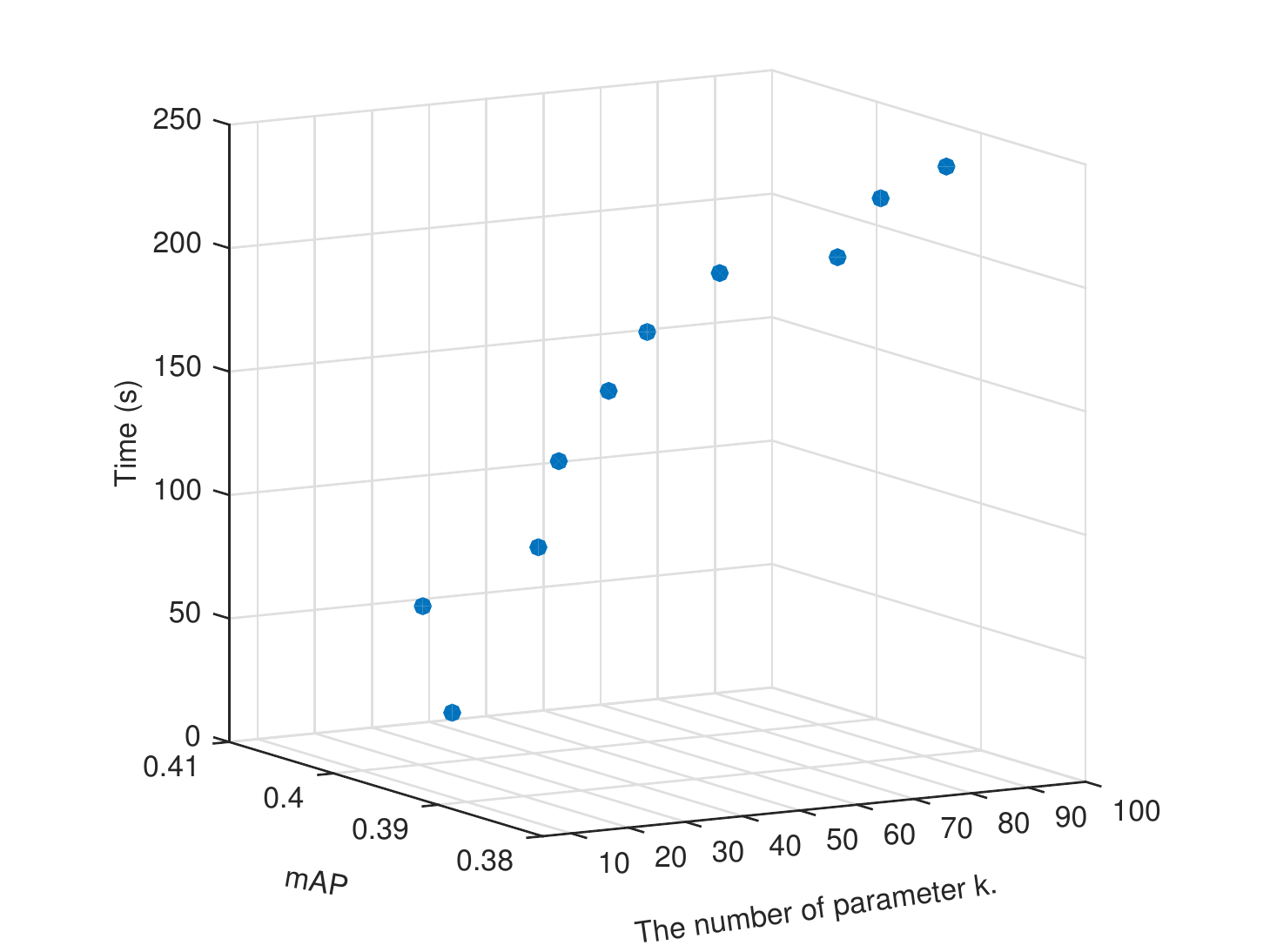}}
}
\end{minipage}

\end{center}
\setlength{\abovecaptionskip}{-8pt}
\caption{The parameter analysis when hash bit is 64.}
\end{figure}

%%%%%%%%%%%%%%%%%%%%%%%%%%%%%%%%%%%%%%%%%%%%%%%%%%%%%%%%%%%%%%%
\section{Conclusion}
In this paper, we proposed a novel unsupervised hashing approach, dubbed Ordinal Constraint Hashing (OCH), for large-scale similarity retrieval.
Unlike most previous unsupervised hashing, the proposed approach exploits the ordinal information between data points, and embeds such relations into a Hamming space.
Firstly, a tensor ordinal graph was proposed to approximate the ordinal relations efficiently.
Then, OCH adopted an ordinal constraint projection scheme to significantly reduce the scale of ordinal graph, which preserves the overall ordinal relation through a small centers set (such as K-means centers).
In optimization, a novel iterative stochastic gradient descent algorithm on Stiefel manifold was developed.
Extensive experiments on three benchmark datasets demonstrated that the proposed OCH approach achieves the best performance in contrast with representative and  state-of-the-art hashing methods.
In our future work, we will further extend the proposed method with deep learning, as well as investigating the possibility of large-scale  binarized optimization in binary code learning.

\section{Acknowledgement}
This work is supported by the National Key RD Program (No. 2016YFB1001503), the Special Fund for Earthquake Research in the Public Interest No.201508025, the Nature Science Foundation of China (No. 61402388, No. 61422210, No. 61373076, and No. 61572410), the CCF-Tencent Open Research Fund, the Open Projects Program of National Laboratory of Pattern Recognition, and the Xiamen Science and Technology Project (No. 3502Z20153003).

\bibliographystyle{aaai}
\bibliography{egbib}

\end{document}